\documentclass[letterpaper, 10 pt, conference]{ieeeconf}  % Comment this line out if you need a4paper

\IEEEoverridecommandlockouts                              % This command is only needed if 
\usepackage{cite}
\usepackage{amsmath,amssymb,amsfonts}
\usepackage{algorithm}
\usepackage{algorithmicx}
\usepackage{graphicx}
\usepackage{textcomp}
\usepackage{graphicx}
\usepackage{textcomp}
\usepackage{xcolor}
\usepackage{dsfont}
\usepackage{subfigure} 
\usepackage[noend]{algpseudocode}

\DeclareMathOperator*{\argmin}{argmin}
\DeclareMathOperator*{\argmax}{argmax}

\title{\LARGE \bf
	An Actor-Critic Method for Simulation-Based Optimization
}

\author{Kuo Li$^{1}$, Qing-Shan Jia$^{1}$, and Jiaqi Yan$^{2}$% <-this % stops a space
	\thanks{$^{1}$K. Li and Q.-S. Jia are with the Center for Intelligent and Networked System (CFINS), Department of Automation, Beijing National Research Center for Information Science and Technology (BNRist), Tsinghua University, Beijing 100084, P.R. China {\tt\small li-k19, jiaqs@mails.tsinghua.edu.cn}}%
	\thanks{$^{2}$Jiaqi Yan is with the Department of Computer Science, Tokyo Institute of Technology, 4259-J2-54, Nagatsuta-cho, Midori-ku, Yokohama 226-8502, Japan
		{\tt\small jyan@sc.dis.titech.ac.jp}}
	\thanks{This work was supported by the NSFC under Grant 62073182 and the 111
		International Collaboration Project (BP2018006).}
}

\begin{document}

	\maketitle
	\thispagestyle{empty}
	\pagestyle{empty}

	%%%%%%%%%%%%%%%%%%%%%%%%%%%%%%%%%%%%%%%%%%%%%%%%%%%%%%%%%%%%%%%%%%%%%%%%%%%%%%%%
	\begin{abstract}
		
		We focus on a simulation-based optimization problem of choosing the best design from the feasible space. Although the simulation model can be queried with finite samples, its internal processing rule cannot be utilized in the optimization process. We formulate the sampling process as a policy searching problem and give a solution from the perspective of Reinforcement Learning (RL). Concretely, Actor-Critic (AC) framework is applied, where the Actor serves as a surrogate model to predict the performance on unknown designs, whereas the actor encodes the sampling policy to be optimized. We design the updating rule and propose two algorithms for the cases where the feasible spaces are continuous and discrete respectively. Some experiments are designed to validate the effectiveness of proposed algorithms, including two toy examples, which intuitively explain the algorithms, and two more complex tasks, i.e., adversarial attack task and RL task, which validate the effectiveness in large-scale problems. The results show that the proposed algorithms can successfully deal with these problems. Especially note that in the RL task, our methods give a new perspective to robot control by treating the task as a simulation model and solving it by optimizing the policy generating process, while existing works commonly optimize the policy itself directly.
		
	\end{abstract}

	%%%%%%%%%%%%%%%%%%%%%%%%%%%%%%%%%%%%%%%%%%%%%%%%%%%%%%%%%%%%%%%%%%%%%%%%%%%%%%%%
	\section{INTRODUCTION}
	
	Simulation-based optimization problems arise in various disciplines such as computer science, economics, and so on. In general, an simulation-based optimization problem consists of maximizing (minimizing) a score (cost) function by choosing appropriate designs from an allowed set, and can be typically formulated as
	\begin{equation}
	\label{equ:op}
	\begin{aligned}
	\argmax\limits_{x}~&~~~f_0(x)\\
	\text{s.t.} ~&f_i(x)\leq 0,~ i=1,2,\cdots,m,
	\end{aligned}
	\end{equation}
	where $x$ is a design from the allowed set $X$ defined by the constraints $f_i(x)\leq 0, ~i=1, 2, \cdots,m$, and $f_0(\cdot)$ is the objective function, which is usually computationally expensive and can be queried only for limited times. The design $x$ can either be a scalar or vector depending on the problem.
	
	There are various works to deal with different cases of \eqref{equ:op}, e.g., convex optimization deals with the case that all of $f_i(\cdot), ~i=1,2,\cdots, m$ and $-f_0(\cdot)$ are convex (\hspace{1pt}\cite{boyd2004convex}), and deep learning deals with the case that $f_0(\cdot)$ is totally or partially calculated by Neural Networks (NNs). However, in both above two cases, the operation rules are designed and utilized during the optimization process, e.g., convex optimization utilizes the gradient or second gradient of $f_0(\cdot)$, and deep learning requires the gradient to all parameters of the NNs. Thus these methods are not suited for simulation-based optimization, where the gradient message cannot be obtained explicitly.
	
	In order to address the issues brought by the ``incomplete'' objective functions, researchers propose to ``reconstruct'' it with a controllable surrogate model, which is usually a predictive model constructed by fitting the existing input-output pairs. Since that a certain deviation between the surrogate model and the true simulation model $f_0(\cdot)$ cannot be avoided, related methods are usually designed to alternate between updating the surrogate model and the sampling distribution. For example, in Bayesian Optimization (BO, \cite{BO}), the Gaussian Process (GP, \cite{GP}) serves as the surrogate model, and the updating process alternates between proposing the best design with the acquisition function constructed with the surrogate model and updating the surrogate model with the latest input-output pair. BO has the advantage of high efficiency, but there are some technical difficulties, e.g., finding the design $x$ with the highest score on the acquisition function is non-trivial, especially in high dimensional cases. The acquisition function is a predictive model, which can only predict the performance on any $x\in \mathcal{X}$, rather than generating the best design $x^*$, which is essential in many real tasks, such as identifying the best parameters for controlling the robots. In these cases, beyond the surrogate model, a generative model must be equipped to identify the best design $x^*$.
	
	The generative model generates a distribution over $\mathcal{X}$, which controls the sampling process. Therefore, the sampling distribution can be bridged to a stochastic control problem (\hspace{1pt}\cite{SBOWSC}), which adjusts the sampling distribution basing on previous information to reach higher expected score on the simulation model $f_0(\cdot)$. 
	The generative model is commonly used in AC-based RL, e.g., DDPG (\hspace{1pt}\cite{lillicrap2015continuous}), TD3 (\hspace{1pt}\cite{fujimoto2018addressing}) and SAC (\hspace{1pt}\cite{haarnoja2018soft}). It is also termed as ``actor'' in AC framework, which generates a distribution over the action space (similar as the design space in simulation-based optimization) for each system state (more details can be found in papers mentioned above).
	
	In this work, we also consider the simulation-based optimization problem, where the operation rule of the simulation model $f(\cdot)$ cannot be utilized in the optimizing process. Similar as BO, a surrogate model will be used to predict the scores on untested designs. In order to overcome the difficulty of finding the global optimal of the acquisition function, which is usually highly non-linear, we also introduce a generative model in our framework. Therefore, our methods inherit the framework of AC in RL area. The framework here is especially similar to that of SAC, which is the state-of-art RL algorithm. The predictive model is termed as critic, which predicts $\hat{y}\approx f_0(x)$ for each $x\in \mathcal{X}$, and the generative model is termed as actor, which will be updated to reach higher expected score on $f_0(\cdot)$. Each iteration starts with generating designs by the actor. These designs will be scored by the simulation model, and the obtained input-output pairs will be used to update the critic. The final step is updating the actor with the latest critic.
	
	The contributions of this paper are mainly as follows: 
	
	1)  Two AC-based algorithms are designed to solve the simulation-based optimization problem, where design space can be either continuous or discrete;
	
	2)  We take two high dimensional tasks, adversarial attack and RL, as examples to show the application on real tasks, and the results show that the proposed algorithms can successfully find satisfying solutions, even in the high dimensional cases;
	
	3) Compared with existing RL methods, e.g., Dynamic Programming (DP, \cite{DP}), Q-learning (\hspace{1pt}\cite{Q-learning}), DDPG, and SAC, which are motivated by optimizing the policy itself, the proposed algorithms offer a new perspective to robot control of reinforcing the policy through optimizing the policy generating process. Besides, the induced methods can be deployed in an off-policy fashion and avoid the problem of sparse and delayed reward.
	
	\section{PRELIMINARY}
	\label{sec:pre}
	Let us begin with introducing the problem of interest and the Actor-Critic (AC) framework extensively used in Reinforcement Learning (RL) technology.
	
	\subsection{Optimization Model}
	The simulation model $f(\cdot)$ takes a scalar (for one dimensional case) or vector (for multi-dimensional case) $x$ as input, and outputs corresponding score $y$, and we aim to find out the best design, i.e.,
	\begin{equation}
	\label{equ:pro}
	\begin{aligned}
	\argmax\limits_{x\in\mathcal{X}}&~f(x),
	\end{aligned}	
	\end{equation}
	where $\mathcal{X}$ is the allowed set.
	
	In real implementations, $f(x)$ could serve as any objective function. For example, it can be a convex optimization problem, if $f(\cdot)$ is a concave function as well as $\mathcal{X}$ is a convex set. $f(\cdot)$ can also be a complex nonlinear function, e.g., Neural Networks (NNs), and the problem comes to find a special design $x$ which receives the maximal response. For another example, in RL tasks, the agent usually intends to find a parameterized policy, which can gather higher accumulated rewards by interacting with the environment. In this case, the environment is a simulation model $f(\cdot)$, and the policy is a design $x$. Then it also comes to the standard form of \eqref{equ:pro}. In Section \ref{sec:app}, we will demonstrate the applications in these tasks.
	
	\subsection{Actor-Critic Framework}
	AC framework is widely used in RL area, which consists of two parts: actor, which generates actions, and critic, which evaluates each action. In RL area, both the actor and critic are typically constructed with NNs, where the actor network encodes a mapping from state $s$ to action $a$, and the critic network maps each $(s,a)$ pair to its predicted accumulated reward. Related work can be found in DDPG (\hspace{1pt}\cite{lillicrap2015continuous}), TD3 (\hspace{1pt}\cite{fujimoto2018addressing}) and SAC (\hspace{1pt}\cite{haarnoja2018soft}).
	
	Typically, the critic is trained by minimizing the residual error of the Bellman equation:
	\begin{equation}
	\begin{aligned}
	L(\phi)=\frac{1}{2}\mathop{\mathds{E}}\left[\left(Q^\phi(s,a)-\left(R(s,a)+\gamma\mathop{\mathds{E}}Q^\phi(s^\prime,a^\prime)\right)\right)^2\right],
	\end{aligned}
	\end{equation}
	where $Q^{\phi}$ is the critic parameterized by $\phi$, $R(s,a)$ is the instantaneous reward, $s^\prime$ is the succeeding state and $a^\prime$ is the next action drawn from current policy. Besides, $\gamma\in[0, 1)$ is the discount factor for stabilizing the training process. $\phi$ can be updated by minimizing $L(\phi)$.
	
	The actor is updated by maximizing the expected accumulated reward:
	\begin{equation}
	\begin{aligned}
	J(\theta)=\mathop{\mathds{E}}\limits_s\left[
	\mathop{\mathds{E}}\limits_{a\sim\pi^\theta(s)}Q^\phi(s,a)
	\right],
	\end{aligned}
	\end{equation}
	where $\pi^\theta$ is the current policy parameterized by $\theta$. Notice that in this work, we choose the stochastic policy instead of the deterministic one for its excellent performance of balancing the exploitation and exploration.
	
	\section{MAIN METHODS}
	
	\label{sec:mm}
	Let us introduce the methodology of solving \eqref{equ:pro} with AC framework. The cases where $\mathcal{X}$ is continuous and discrete will be separately considered.
	
	\begin{figure}
		\centering
		\subfigure[Actor for continuous design space]{
			\includegraphics[scale=0.5]{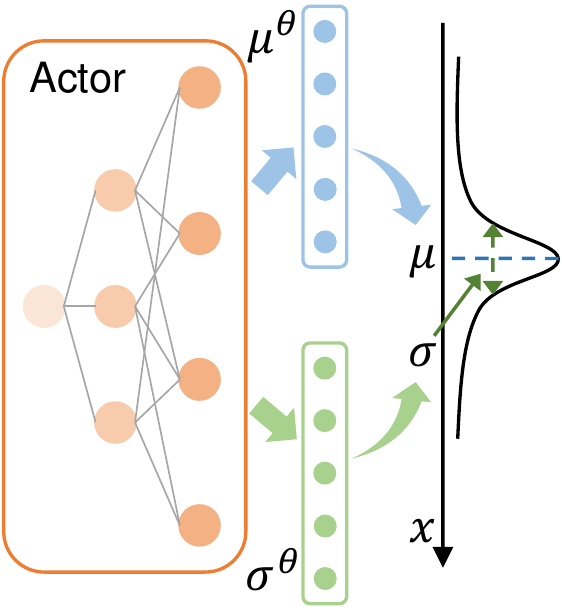}\label{fig:continuous}}
		\hspace{0pt}
		\subfigure[Actor for discrete design space]{
			\includegraphics[scale=0.5]{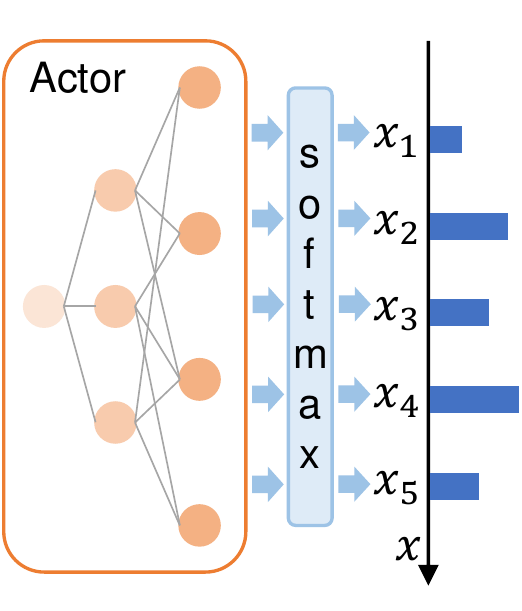}\label{fig:discrete}}
		\hspace{0pt}
		\subfigure[Critic]{
			\includegraphics[scale=0.5]{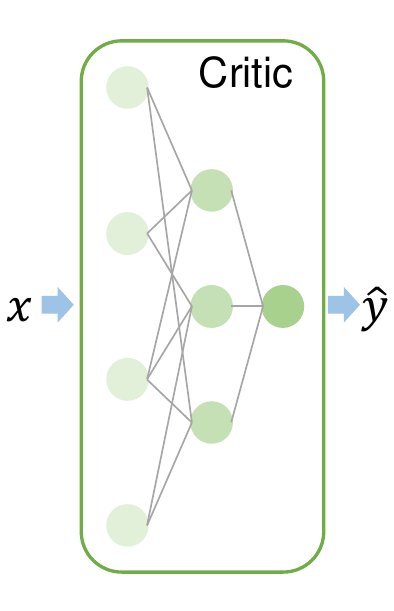}\label{fig:critic}}
		\caption{Illustration of the network structures. Fig. \ref{fig:continuous} and Fig. \ref{fig:discrete} are the network structures of the actor for continuous and discrete design space respectively. Fig. \ref{fig:critic} is the network structure of the critic.}
		\label{fig:NN}
	\end{figure}
	
	We use NNs to construct the actor and critic (Fig. \ref{fig:NN}). The actor takes a random noise (or constant) as input to capture the multi-modal (or single-modal) distributions, and the critic takes the design $x$ as input and outputs a predicted score.
	
	The actor and critic are updated iteratively. Each iteration starts with generating a feasible design $x$ and getting its score $y=f(x)$ by querying the simulation model $f(\cdot)$. The $(x, y)$ pair will be stored in a replay buffer $\mathcal{D}$. Then a mini-batch $\mathcal{B}_q$ will be sampled from $\mathcal{D}$ to update the critic by minimizing the loss function:
	\begin{equation}
	L(\phi) = \frac{1}{2}\sum\limits_{(x,y)\in \mathcal{B}_q}\left(Q^\phi(x)-y\right)^2,
	\end{equation}
	with one step of gradient descent:
	\begin{equation}
	\label{equ:lq}
	\nabla_\phi L(\phi) = \sum\limits_{(x,y)\in \mathcal{B}_q}\left(Q^\phi(x)-y\right)\nabla_\phi Q^\phi(x).
	\end{equation}
	Finally, the actor will be updated with the latest critic.
	
	A stochastic actor is used to generate designs, and in each iteration, the sampling distribution will be updated to improve the expected score. In order to reduce the risk of falling into local optimal, we use entropy regularization to reach the trade-off between exploitation and exploration. The updating rule of the actor will be studied for continuous and discrete design space respectively.
	
	\subsection{Continuous Design Space}
	\label{sec:cvr}
	We firstly introduce the case where $\mathcal{X}$ is continuous. For the ease of illustration, this paper focuses on the limited space $\mathcal{X}=\left\{\forall x~|~||x||_\infty<1\right\}$. However, we note that the results can be easily extended to any continuous limited space through linear mapping.
	
	Squashed Gaussian policy (similar as SAC) is used to generate designs, as shown in Fig. \ref{fig:continuous}. Firstly, the actor network outputs two vectors, $\mu^\theta$ and $\sigma^\theta$, where $\theta$ is the parameter of the actor network. Then $x$ is generated by
	\begin{equation}
	x=g^\theta(\xi)=tanh(\mu^\theta+\sigma^\theta\odot\xi),
	\end{equation}
	where $\odot$ represents the element-wise multiplication. $\mu^\theta$, $\sigma^\theta$ and $\xi$ have the same dimension with $x$, and $\xi$ is a random noise taken from multi-dimensional standard
	Gaussian distribution $\xi\sim\mathcal{N}(0, I)$. The $tanh$ function converts $\mathcal{N}(\mu^\theta, diag\left(\sigma^\theta\right))$ to another distribution whose value range is $(-1, 1)$, and its probability density function can be calculated by
	\begin{equation}
	\begin{aligned}
	\hbar(x|\theta) =& h(\xi|\theta)\left|\det(\frac{dg^\theta(\xi)}{d\xi})\right|^{-1}
	\end{aligned}
	\end{equation}
	at $\xi=(g^\theta)^{(-1)}(x)$, where $h(\cdot|\theta)$ is the probability density function of $\mathcal{N}(\mu^\theta, diag\left(\sigma^\theta\right))$, and $(g^\theta)^{(-1)}(\cdot)$ is the inverse function of $g^\theta(\cdot)$.
	
	The sampling distribution will be updated to reach higher expected score
	\begin{equation}
	\label{equ:j}
	J(\theta)=\mathop{\mathds{E}}\limits_{x\sim\hbar(\cdot|\theta)}Q^\phi(x) + \alpha H(\hbar(\cdot|\theta)),
	\end{equation}
	where $H(\hbar(\cdot|\theta))$ is the entropy regularization for balancing exploitation and exploration, and can be calculated by
	\begin{equation}
	\label{equ:ent}
	H(\hbar(\cdot|\theta)) = \mathop{\mathds{E}}\limits_{x\sim\hbar(\cdot|\theta)}[-\log\hbar(x|\theta)].
	\end{equation}
	Therefore, $J(\theta)$ can be rewritten as
	\begin{equation}
	\begin{aligned}
	J(\theta)=&\mathop{\mathds{E}}\limits_{x\sim\hbar(\cdot|\theta)}\left[Q^\phi(x)-\alpha \log\hbar(x|\theta)\right]\\
	=&\mathop{\mathds{E}}\limits_{\xi\sim\mathcal{N}(0,I)}\left[Q^\phi(g^\theta(\xi))-\alpha \log\hbar(g^\theta(\xi)|\theta)\right]
	\end{aligned}
	\end{equation}
	by substituting \eqref{equ:ent} to \eqref{equ:j}, and its gradient with respect to $\theta$ can be derived by:
	\begin{equation}
	\label{equ:nj}
	\begin{aligned}
	\nabla_\theta J(\theta)= &\mathop{\mathds{E}}\limits_{\xi\sim\mathcal{N}(0, I)}\left[\nabla_{g^\theta(\xi)}\left( Q^\phi(g^\theta(\xi))\right.\right.\\
	&\left.\left.-\alpha\log\hbar(g^\theta(\xi)|\theta)\right)\nabla_\theta g^\theta(\xi)\right],
	\end{aligned}
	\end{equation}
	which can be estimated by a batch of samples $\mathcal{B}_a$ taken from $\mathcal{N}(0,I)$:
	\begin{equation}
	\label{equ:eg}
	\begin{aligned}
	\nabla_\theta J(\theta)=& \frac{1}{N}\sum\limits_{\xi\in\mathcal{B}_a}\left[\nabla_{g^\theta(\xi)}\left( Q^\phi(g^\theta(\xi))\right.\right.\\
	&\left.\left.-\alpha\log\hbar(g^\theta(\xi)|\theta)\right)\nabla_\theta g^\theta(\xi)\right],
	\end{aligned}
	\end{equation}
	where $N$ is the size of $\mathcal{B}_a$. Finally, $\theta$ can be updated with \eqref{equ:eg} through one step of gradient ascent.
	
	We conclude the optimization process in Algorithm \ref{alg:cvr}. In each iteration, only a single design will be taken to be validated with the optimization model, and the data pair will be stored for future use. In the end, the best design in the replay buffer will be chosen as the final result.
	
	\begin{algorithm}
		\caption{Optimization with continuous design space\label{alg:cvr}}
		\begin{algorithmic}[1]
			\For{$episode=1:M$}
			\State The actor generates a design 
			\begin{equation}
			x=g^\theta(\xi)=tanh(\mu^\theta+\sigma^\theta\odot\xi), ~\xi\sim\mathcal{N}(0, I).
			\end{equation}
			\State The simulation model gives the score $y=f(x)$.
			\State Store $(x, y)$ pair to $\mathcal{D}$.
			\For{$round=1:K$}
			\State Sample a mini-batch $\mathcal{B}_q$ from $\mathcal{D}$.
			\State Update the critic with gradient estimated by \eqref{equ:lq}.
			\EndFor
			\For{$round=1:K$}
			\State Sample a mini-batch $\mathcal{B}_a$ from $\mathcal{N}(0, I)$.
			\State Update the actor with gradient estimated by \eqref{equ:eg}.
			\EndFor
			\EndFor
			\State \Return the best design stored in $\mathcal{D}$.
		\end{algorithmic}
	\end{algorithm}
	
	\subsection{Discrete Design Space}
	
	Here we consider the case that there are only finite feasible designs in $\mathcal{X}$. In this case, the network structure of the actor will be slightly modified to output a discrete distribution over the design space. As illustrated in \ref{fig:discrete}, the number of the output channels equals the size of $\mathcal{X}$, and each channel corresponds to a feasible design. After applying the softmax function, a distribution over $\mathcal{X}$ will be obtained:
	\begin{equation}
	\label{equ:pp}
	P^\theta(x_i)=\frac{\exp\left(\mu^\theta(x_i)\right)}{\sum\limits_{x_j\in \mathcal{X}}\exp\left(\mu^\theta(x_j)\right)}, ~\forall x_i\in \mathcal{X},
	\end{equation} 
	where by slightly abusing of notation, $\mu^\theta(x_i)$ here denotes the output of the channel corresponding to $x_i$. Then we update the actor by maximizing the expected score:
	\begin{equation}
	\label{equ:obj}
	\begin{aligned}
	J(\theta) =& \mathop{\mathds{E}}\limits_{x\sim P^\theta}Q^\phi(x)+ \alpha H(P^\theta(\cdot))\\
	=&\sum\limits_{x\in \mathcal{X}}P^\theta(x)\left[Q^\phi(x)-\alpha\log P^\theta(x)\right],
	\end{aligned}	
	\end{equation}
	where the entropy regularization is also introduced as in Section \ref{sec:cvr}. Then, the gradient of $J$ with respect to $\theta$ can be calculated by:
	\begin{equation}
	\label{equ:esg}
	\begin{aligned}
	\nabla_{\theta} J(\theta) = \sum\limits_{x\in \mathcal{X}}\nabla_\theta P^\theta(x)\left[Q^\phi(x)-\alpha \log P^\theta(x)-\alpha\right],
	\end{aligned}
	\end{equation}
	and the actor can be updated with one step of stochastic gradient ascent.
	
	\begin{algorithm}
		\caption{Optimization with discrete design space\label{alg:dvr}}
		\begin{algorithmic}[1]
			\For{$episode=1:M$}
			\State The actor generates a design $x\sim P^\theta(\cdot)$.	
			\State The simulation model gives the score $y=f(x)$.
			\State Store $(x, y)$ pair to $\mathcal{D}$.
			\For{$round=1:K$}
			\State Sample a mini-batch $\mathcal{B}_q$ from $\mathcal{D}$.
			\State Update the critic with gradient estimated by \eqref{equ:lq}.
			\begin{equation}
			\nabla_\phi L(\phi) = \sum\limits_{(x,y)\in \mathcal{B}_q}\left(Q^\phi(x)-y\right)\nabla_\phi Q^\phi(x).
			\end{equation}
			\EndFor
			\For{$round=1:K$}
			\State The actor generates the distribution $P^\theta(\cdot)$.
			\State The critic predicts score for each design $Q^\phi(\cdot)$.
			\State Update the actor with gradient estimated by \eqref{equ:esg}.
			\EndFor
			\EndFor
			\State \Return the best design stored in $\mathcal{D}$.
		\end{algorithmic}
	\end{algorithm}
	
	We conclude the optimization process in Algorithm \ref{alg:dvr}. It is worth noting that in each update of the actor, we attempt to maximize the objective function \eqref{equ:obj}, whose optimal solution is the energy-based policy (\hspace{1pt}\cite{haarnoja2017reinforcement}):
	\begin{equation}
	\label{equ:ps}
	P^*(x_i)=\frac{\exp\left(\frac{Q^\phi(x_i)}{\alpha}\right)}{\sum\limits_{x_j\in \mathcal{X}}\exp\left(\frac{Q^\phi(x_j)}{\alpha}\right)}, ~~\forall x_i \in \mathcal{X}.
	\end{equation}
	There is a simple proof of \eqref{equ:ps} by constructing following optimization problem:
	\begin{equation}
	\label{equ:opr}
	\begin{aligned}
	\argmin\limits_{P(x)}~&-\sum\limits_{x\in \mathcal{X}}P(x)\left[Q^\phi(x)-\alpha\log P(x)\right]\\
	\text{s.t.} ~&~~~~~~~-P(x) < 0, ~\forall x \in \mathcal{X}\\
	&\sum\limits_{x\in \mathcal{X}} P(x) - 1 =0.
	\end{aligned}
	\end{equation}
	It is easy to verify that \eqref{equ:opr} is a convex problem and satisfies the Slater condition (\hspace{1pt}\cite{boyd2004convex}), which guarantees that $P^*(x)$ is the unique solution of the KKT conditions:
	\begin{subequations}
		\label{equ:kkt}
		\begin{align}
		P^*(x) &> 0, &\forall& x\in \mathcal{X}\label{equ:kkt1}\\
		\sum\limits_{x\in \mathcal{X}}P^*(x) &= 1&&\label{equ:kkt2}\\
		\lambda^*(x)&\geq 0, &\forall& x\in \mathcal{X}\label{equ:kkt3}\\
		\lambda^*(x)P^*(x) &= 0, &\forall& x \in \mathcal{X}\label{equ:kkt4}\\
		-Q^\phi(x)+\alpha \log P^*(x)&~&~&\notag\\
		+\alpha -\lambda^*(x)+\nu^*&=0, &\forall& x \in \mathcal{X}\label{equ:kkt5},
		\end{align}
	\end{subequations}
	where $\lambda^*$ and $\nu^*$ are the optimal value of dual variables. Combining \eqref{equ:kkt1}, \eqref{equ:kkt3} and \eqref{equ:kkt4}, we can get $\lambda^*(x)=0,\forall x\in \mathcal{X}$. Then, subtracting it into \eqref{equ:kkt5}, we get
	\begin{equation}
	\label{equ:logp}
	\log P^*(x) = \frac{Q^\phi(x)}{\alpha} + c, ~\forall x \in \mathcal{X},
	\end{equation}
	where $c=-1-\frac{\nu^*}{\alpha}$. By subtracting \eqref{equ:logp} into \eqref{equ:kkt2}, we have
	\begin{equation}
	\label{equ:c}
	c=-\log \sum\limits_{x\in \mathcal{X}}\exp \left(\frac{Q^\phi(x)}{\alpha}\right).
	\end{equation}
	Finally, according to \eqref{equ:logp} and \eqref{equ:c}, the optimal solution \eqref{equ:ps} can be obtained. In Section \ref{sec:app}, we will show that the learned distribution is very close to the optimal distribution $P^*$.
	
	So far, we have introduced the methods to solve the simulation-based optimization problem \ref{equ:pro} with AC framework, and proposed algorithms for continuous and discrete design space respectively. In the next section, we will design some experiments to validate them.
	
	\section{APPLICATIONS}
	\label{sec:app}
	In this section, we will start from a toy example, with which some intuitive interpretation of Algorithm \ref{alg:cvr} and \ref{alg:dvr} will be given. Then we design some complex tasks, e.g., adversarial attack and reinforcement learning, to show the effectiveness of the proposed algorithms in large-scale systems.
	
	Firstly, we choose the probability density function of a Gaussian Mixture Model (GMM) as the toy example:
	\begin{equation}
	\label{equ:bm}
	\begin{aligned}
	f(x) =& w_1 \times \frac{1}{\sqrt{2\pi}\sigma_1}\exp\left(-\frac{(x-\mu_1)^2}{2\sigma_1^2}\right)\\
	&+w_2 \times \frac{1}{\sqrt{2\pi}\sigma_2}\exp\left(-\frac{(x-\mu_2)^2}{2\sigma_2^2}\right), ~x\in(-1,1),
	\end{aligned}
	\end{equation}
	and the parameters are set as $w_1=0.51$, $w_2=0.49$, $\mu_1 = -0.7$, $\mu_2=0.7$, $\sigma_1 = 0.6$, $\sigma_2=0.6$.
	
	\begin{figure*}
		\centering
		\subfigure[$\alpha = 10^{-1}$]{
			\includegraphics[scale=0.29]{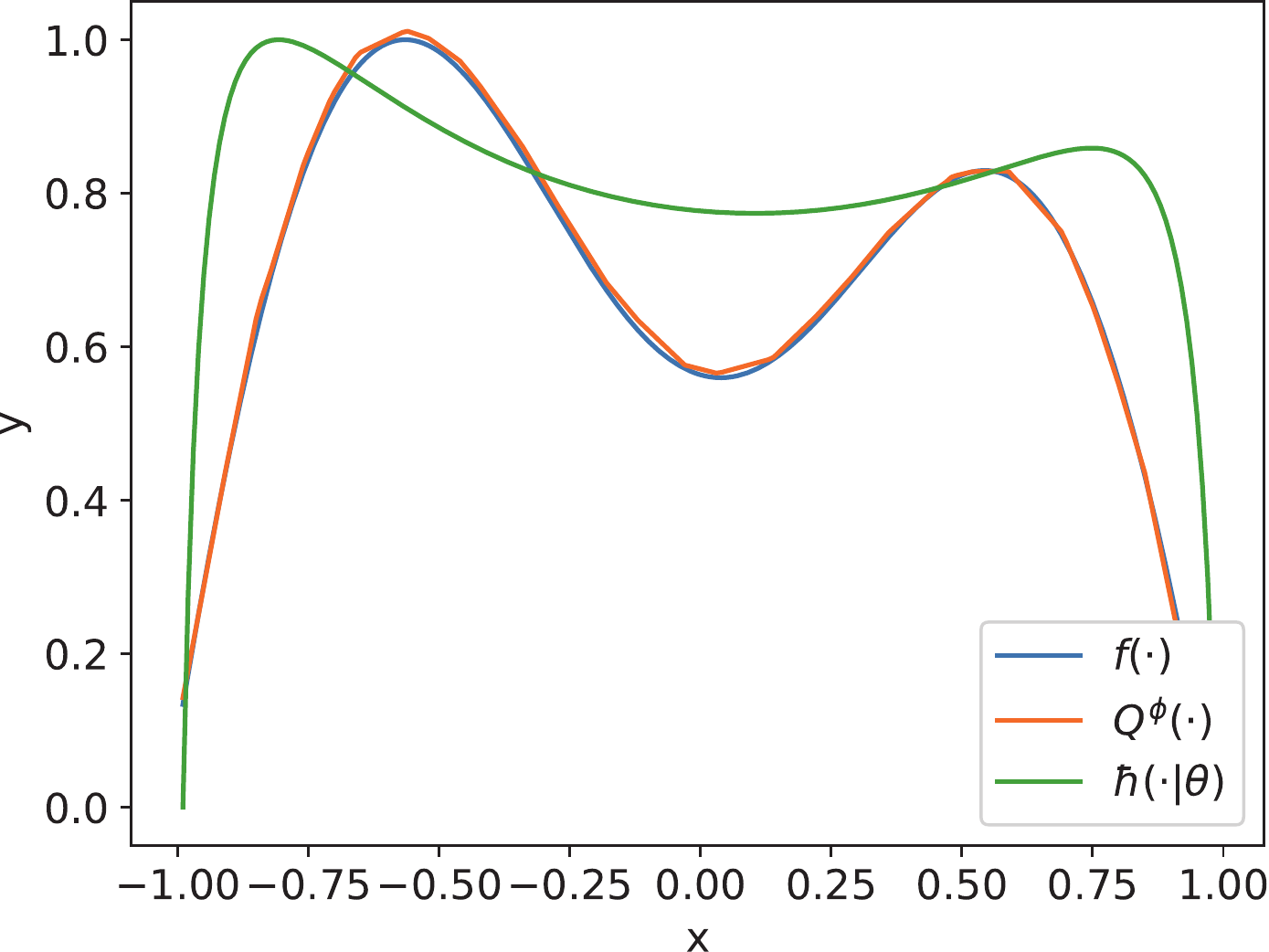}\label{fig:alpha01}}
		\hspace{0pt}
		\subfigure[$\alpha = 10^{-2}$]{
			\includegraphics[scale=0.29]{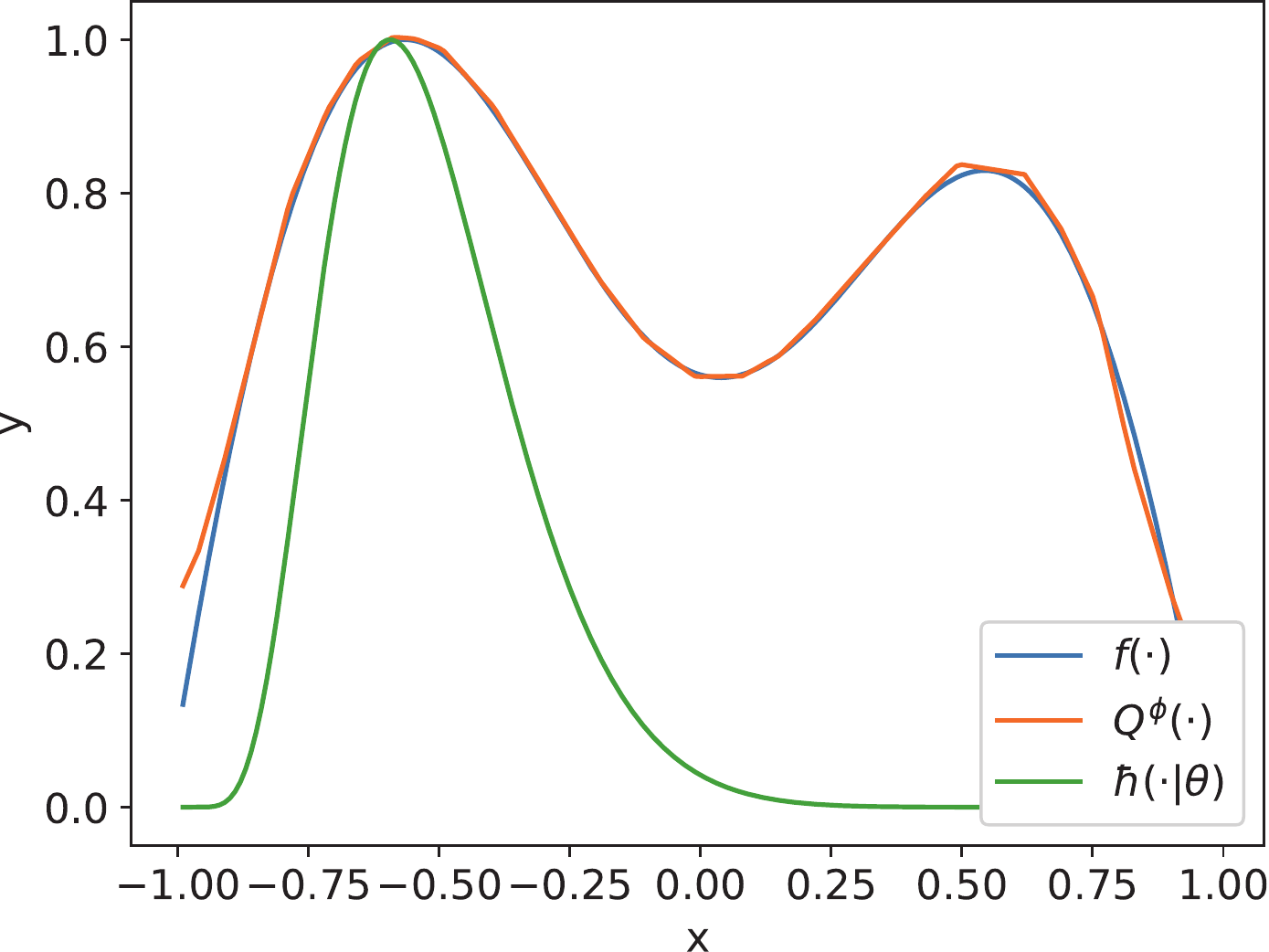}\label{fig:alpha001}}
		\hspace{0pt}
		\subfigure[$\alpha = 10^{-3}$]{
			\includegraphics[scale=0.29]{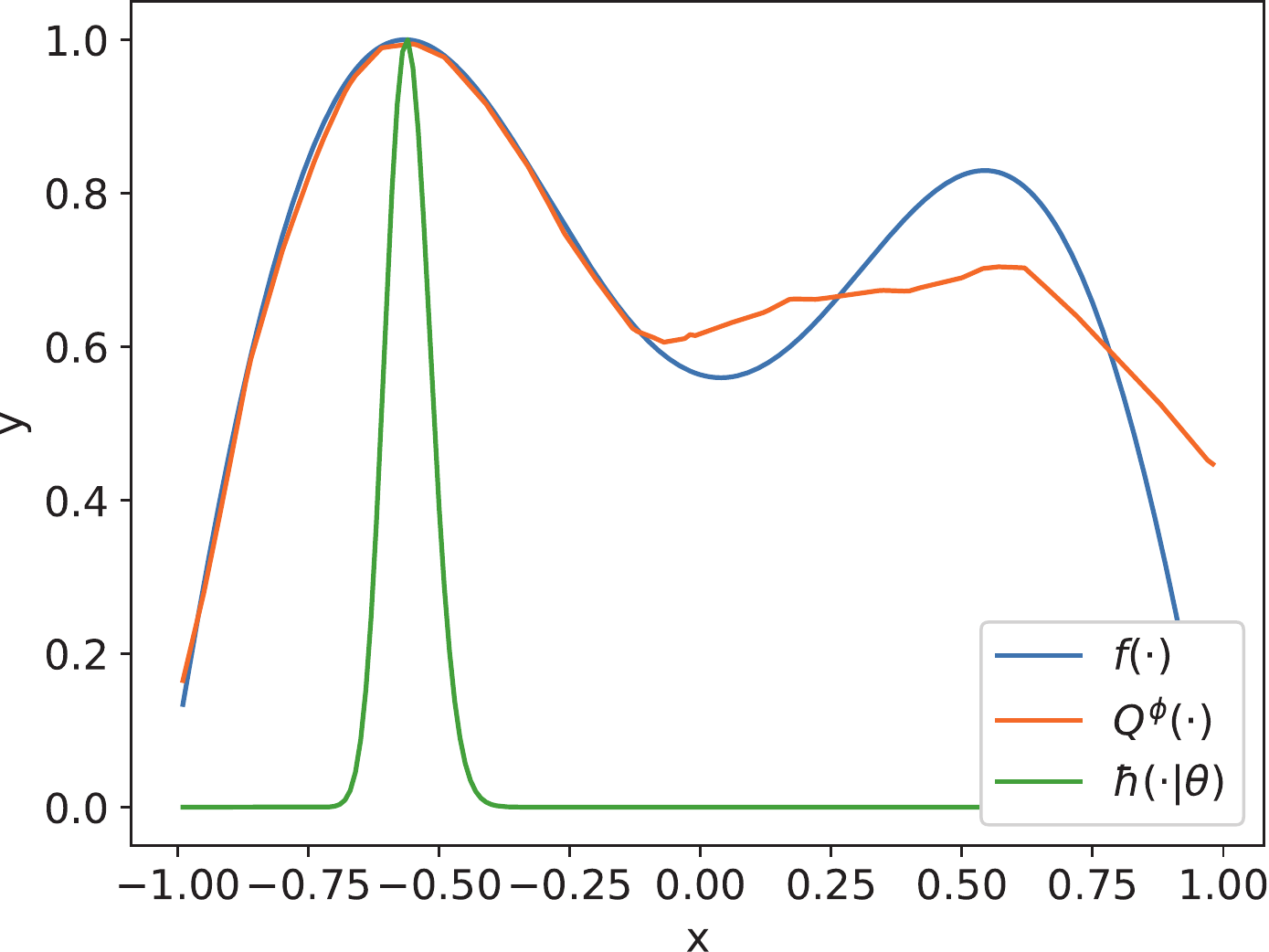}\label{fig:alpha0001}}
		\hspace{0pt}
		\subfigure[$Deterministic$]{
			\includegraphics[scale=0.29]{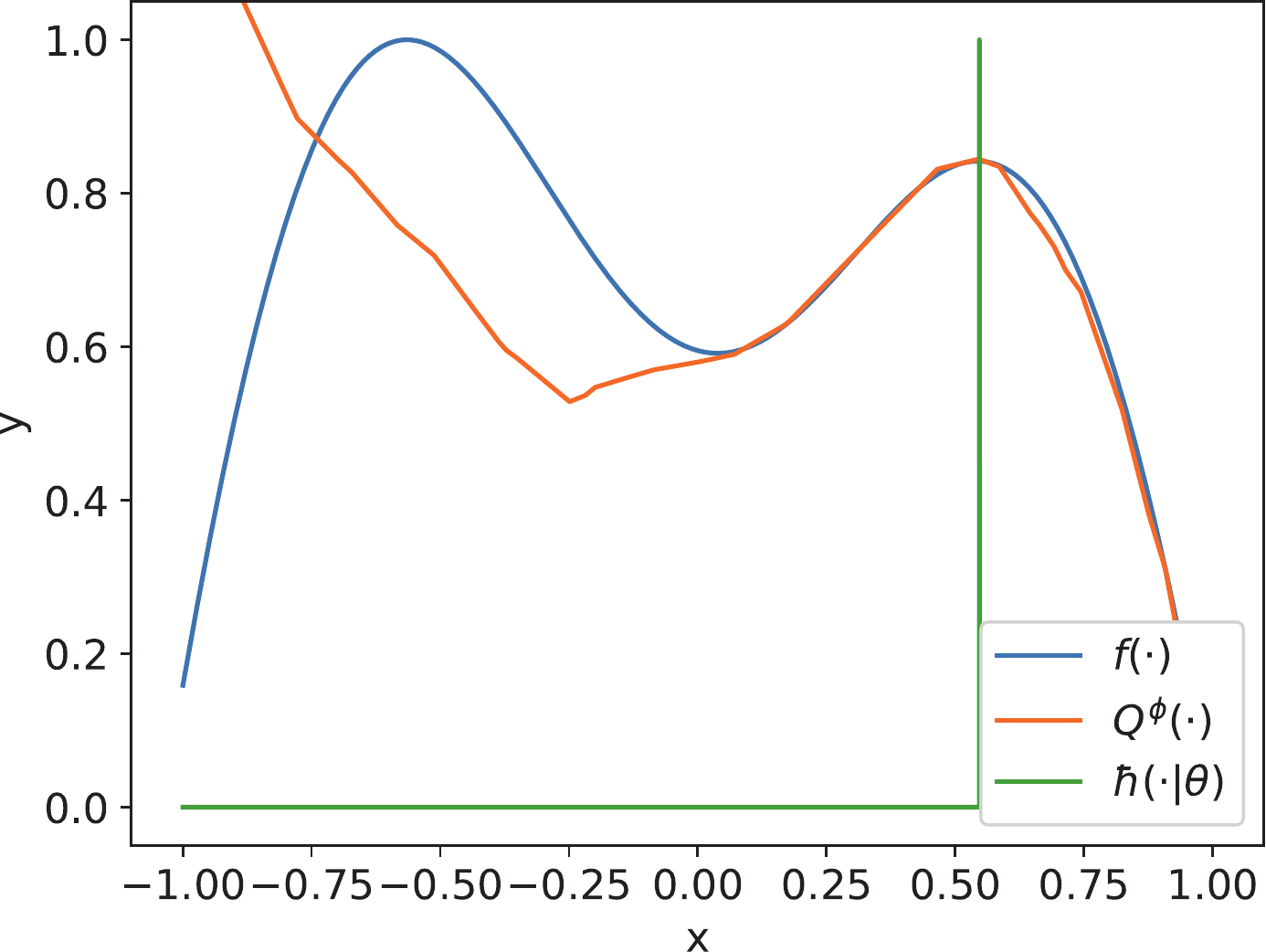}\label{fig:alpha0}}
		\caption{Results of Algorithm \ref{alg:cvr}. We choose $\alpha=10^{-1}$, $\alpha=10^{-2}$ and $\alpha=10^{-3}$ in the experiments, and the results are shown in Fig. \ref{fig:alpha01}, \ref{fig:alpha001} and Fig. \ref{fig:alpha0001}, respectively. In each figure, the blue line is the ground truth of the simulation model $f(\cdot)$. The orange line is the value predicted by the critic $Q^\phi(\cdot)$. The green line is the sampling distribution $\hbar(\cdot|\theta)$ generated by the actor. The curves are normalized to $[0,1]$ for the convenience of comparison. In order to show the effect of the entropy regularization, we replace the stochastic actor with a deterministic one, and the result is shown in Fig. \ref{fig:alpha0}.}
		\label{fig:tc}
	\end{figure*}
	
	We launch Algorithm \ref{alg:cvr} and the results are shown in Fig. \ref{fig:tc}. It can be found that the sampling distribution gradually converges to the optimal area as $\alpha$ decreases, which means by setting an appropriate $\alpha$ to controls the strength of the entropy regularization, satisfying designs can be drawn from the final actor. To analysis the effect of entropy regularization, $\alpha$ is set as $10^{-1}$, $10^{-2}$, and $10^{-3}$ respectively. Compared with smaller $\alpha$ in Fig. \ref{fig:alpha001} and Fig. \ref{fig:alpha0001}, the actor in Fig. \ref{fig:alpha01} generates almost an uniform distribution, which means that the entropy regularization has a greater impact on the objective in \eqref{equ:j}. Besides, since the samples distribute more evenly over the allowed set, the predicted scores are close to the ground truth almost everywhere. With $\alpha$ decreasing, the sampling distribution narrows to the optimal area (Fig. \ref{fig:alpha001} and Fig. \ref{fig:alpha0001}), which means that the actor can generate satisfying designs. However, the samples concentrate in some areas, which leads to worse performance of the critic in other areas, as illustrated in Fig. \ref{fig:alpha0001}. We also test the performance of a deterministic actor, as in DDPG (\hspace{1pt}\cite{lillicrap2015continuous}), where the actor only generates a deterministic design, rather than a distribution over the design space. The final design falls into the local optimal more easily, as illustrated in Fig. \ref{fig:alpha0}. In real implementation, we usually set a larger $\alpha$ in the initial phase, and decrease it step by step. In this way, the sampling distribution can converge to the neighbor area of the global optimal, while avoiding falling into local optimal.
	
	\begin{figure*}
		\centering
		\subfigure[$\alpha = 10^{-1}$]{
			\includegraphics[scale=0.29]{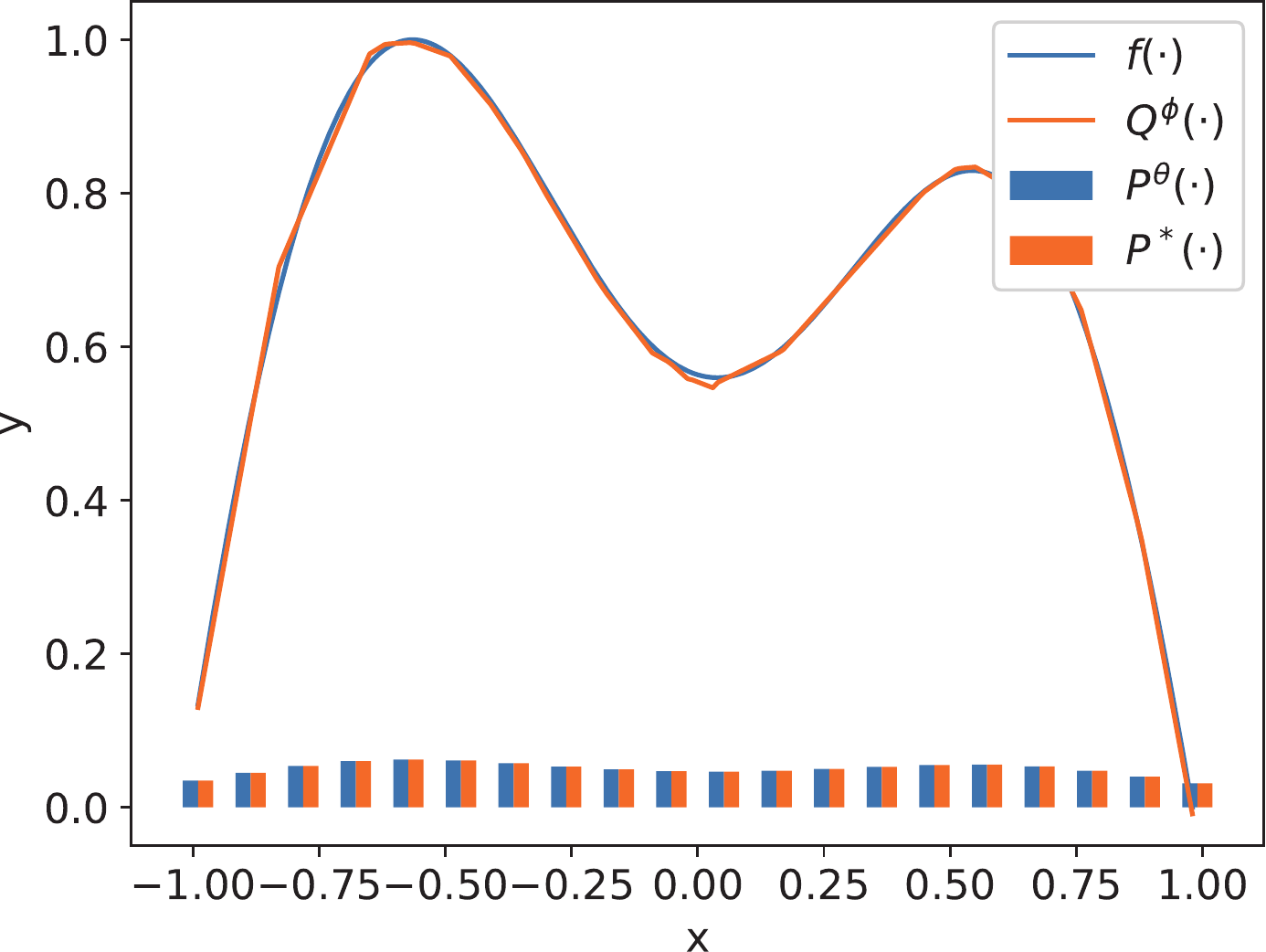}\label{fig:dalpha01}}
		\hspace{0pt}
		\subfigure[$\alpha = 10^{-2}$]{
			\includegraphics[scale=0.29]{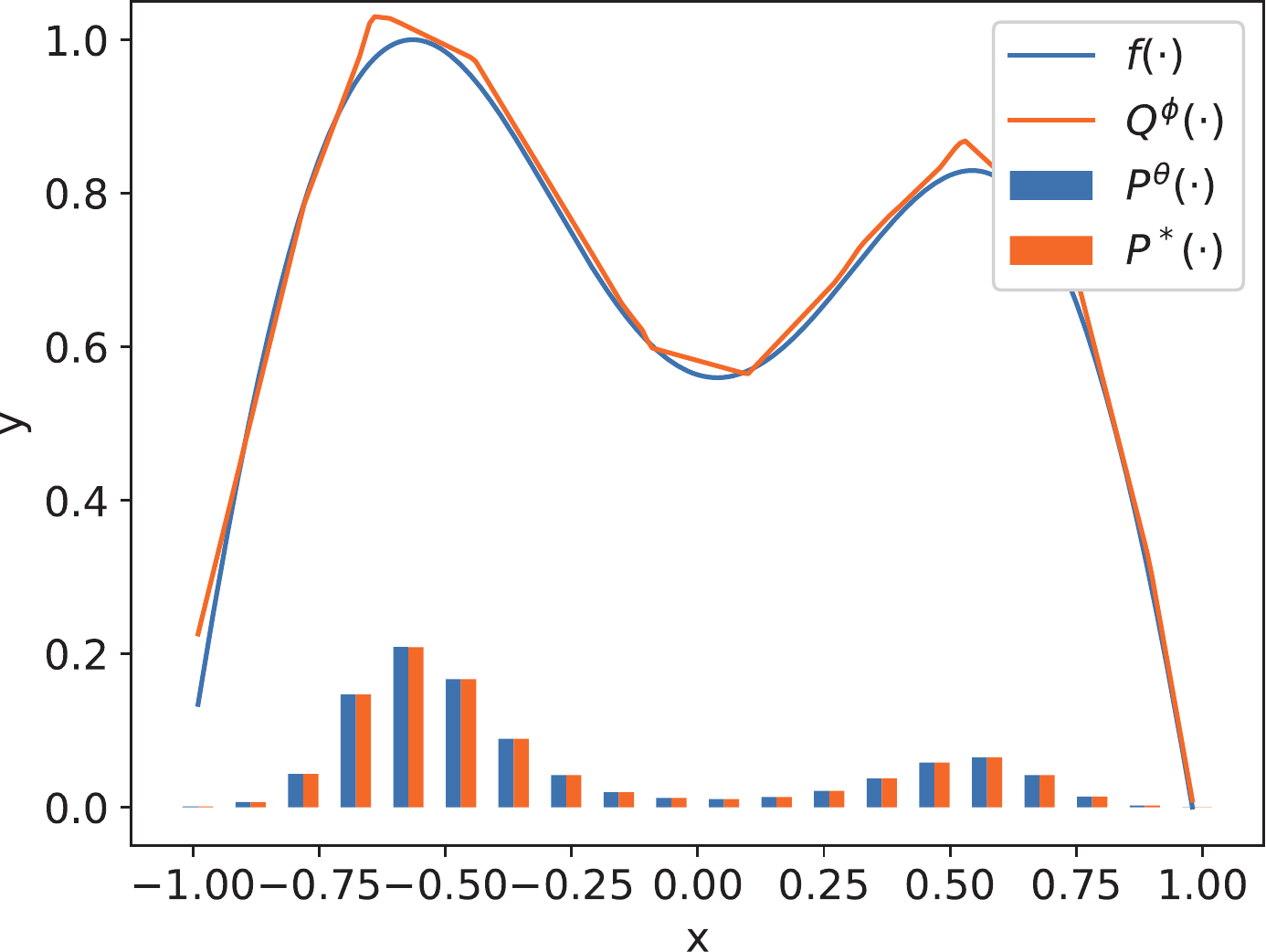}\label{fig:dalpha001}}
		\hspace{0pt}
		\subfigure[$\alpha = 10^{-3}$]{
			\includegraphics[scale=0.29]{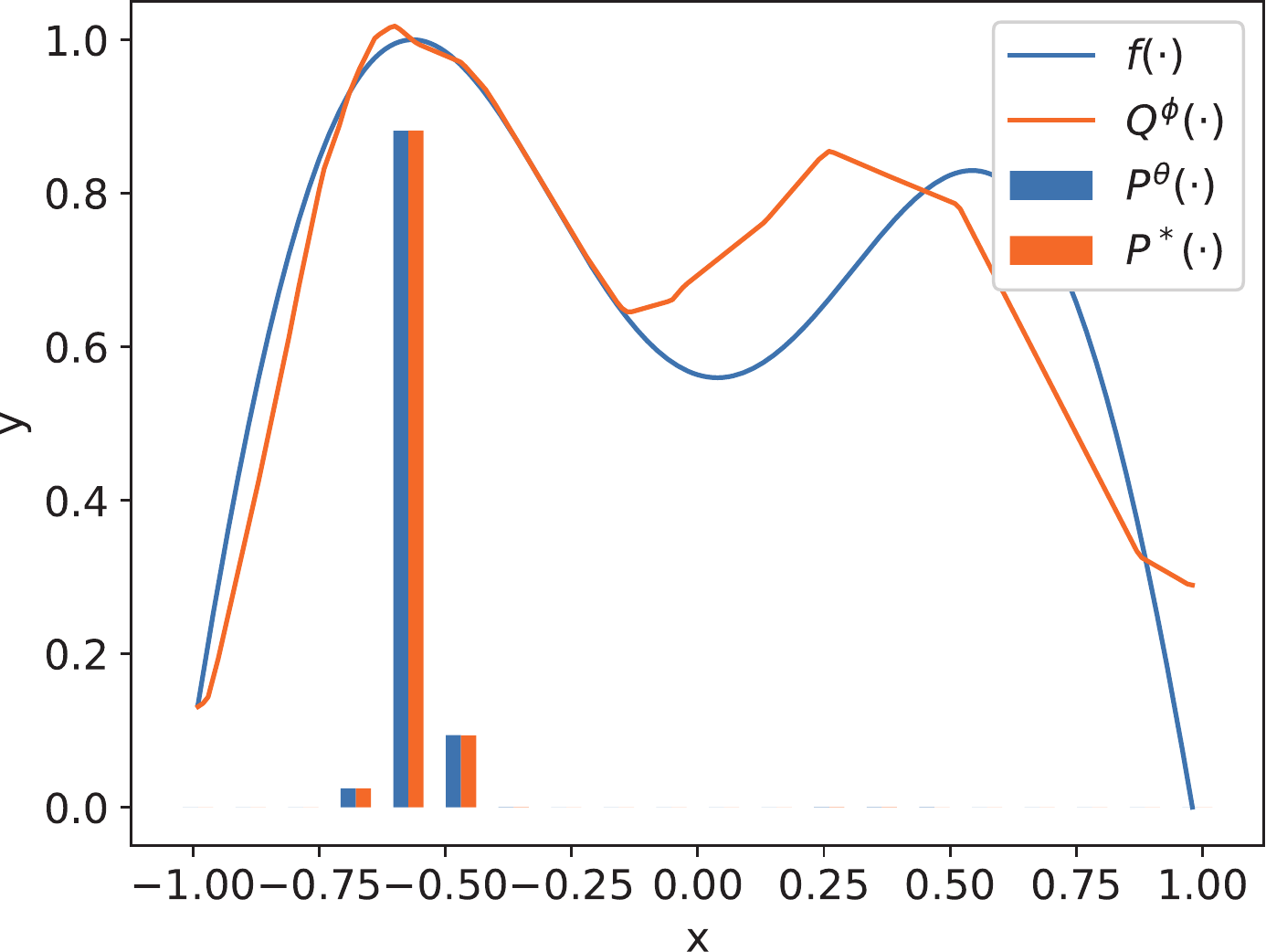}\label{fig:dalpha0001}}
		\hspace{0pt}
		\subfigure[$\alpha=10^{-3}$]{
			\includegraphics[scale=0.29]{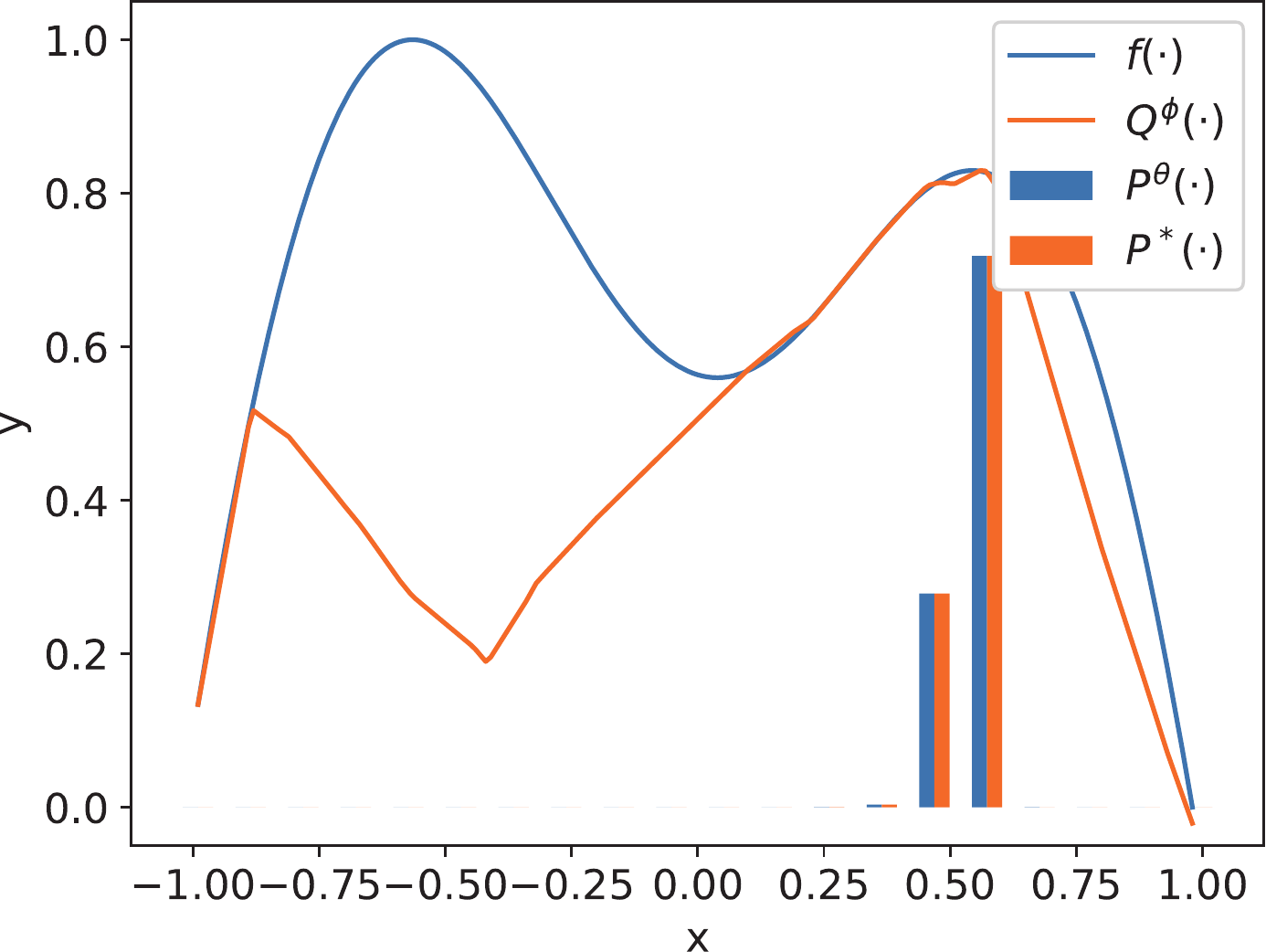}\label{fig:dalpha00010}}
		\caption{Results of Algorithm \ref{alg:dvr}. We choose $\alpha=10^{-1}$, $\alpha=10^{-2}$ and $\alpha=10^{-3}$ in the experiments. As in Fig. \ref{fig:tc}, the blue line is the ground truth of the simulation model $f(\cdot)$ and the orange line is the value predicted by the critic $Q^\phi(\cdot)$. We use histograms to illustrate the sampling distribution among the discrete $\mathcal{X}$. The blue bars denote the sampling distribution $P^\theta(\cdot)$ generated by the actor and the orange bars denote the optimal distribution $P^*(\cdot)$.}
		\label{fig:td}
	\end{figure*}
	
	The simulation model \eqref{equ:bm} is modified to test Algorithm \ref{alg:dvr} by converting the design space $\mathcal{X}$ to a discrete set. The results are shown in Fig. \ref{fig:td}. It can be found that by decreasing the strength of entropy regularization, the sampling distribution gradually converges to the optimal area. We also compare the effect of the entropy regularization by setting $\alpha$ as $10^{-1}$, $10^{-2}$, and $\alpha=10^{-3}$. When a larger $\alpha$ is chosen, e.g., $\alpha=10^{-1}$ in Fig. \ref{fig:dalpha01}, the actor generates nearly uniform distribution. When decreasing it, the learned distribution gradually converges to designs with higher scores, as in Fig. \ref{fig:dalpha001} and Fig. \ref{fig:dalpha0001}. However, if $\alpha$ is set too small, the actor may fall into local optimal, as illustrated in Fig. 
	\ref{fig:dalpha00010}. Therefore, in real implementation, we also set a bigger $\alpha$ in the initial phase and decrease it step by step as in the continuous case. Furthermore, it can be found that the distribution $P^\theta(\cdot)$ generated by the actor is very close to the optimal one $P^*(\cdot)$, which validates the theoretical analysis.
	
	After the toy examples, we design more complex tasks to validate the proposed algorithms.
	
	The first task is abstracted from adversarial attack tasks. In these tasks, we need to generate samples to ``cheat'' some NNs, which are trained to recognize the objects in the input pictures. In this experiment, we train a classifier to recognize digital numbers with the public data set MNIST. The classifier  $D(\cdot)$  takes an image $x^\prime$ as input and generates a distribution over $Y=\{0,1,2,\cdots,9\}$, which denotes the confidence of classifying the digital image to corresponding value. For the convenience of later representation, the distribution is denoted as $D(x^\prime)\in R^{10}$ and $D(x^\prime, i), i\in Y$ denotes the confidence of classifying the image $x^\prime \in R^{W\times H}$ into $i$, where by slight abuse of notations, here the $W$ and $H$ are the size of images. In the adversarial attack tasks, we need to generate some fake images to enhance the data set and train a more robust classifier. 
	
	\begin{figure}
		\centering
		\subfigure[Learning curve of the training process. The curve is drawn by averaging the confidences on $300$ samples generated by the actor after each episode.]{
			\includegraphics[scale=0.3]{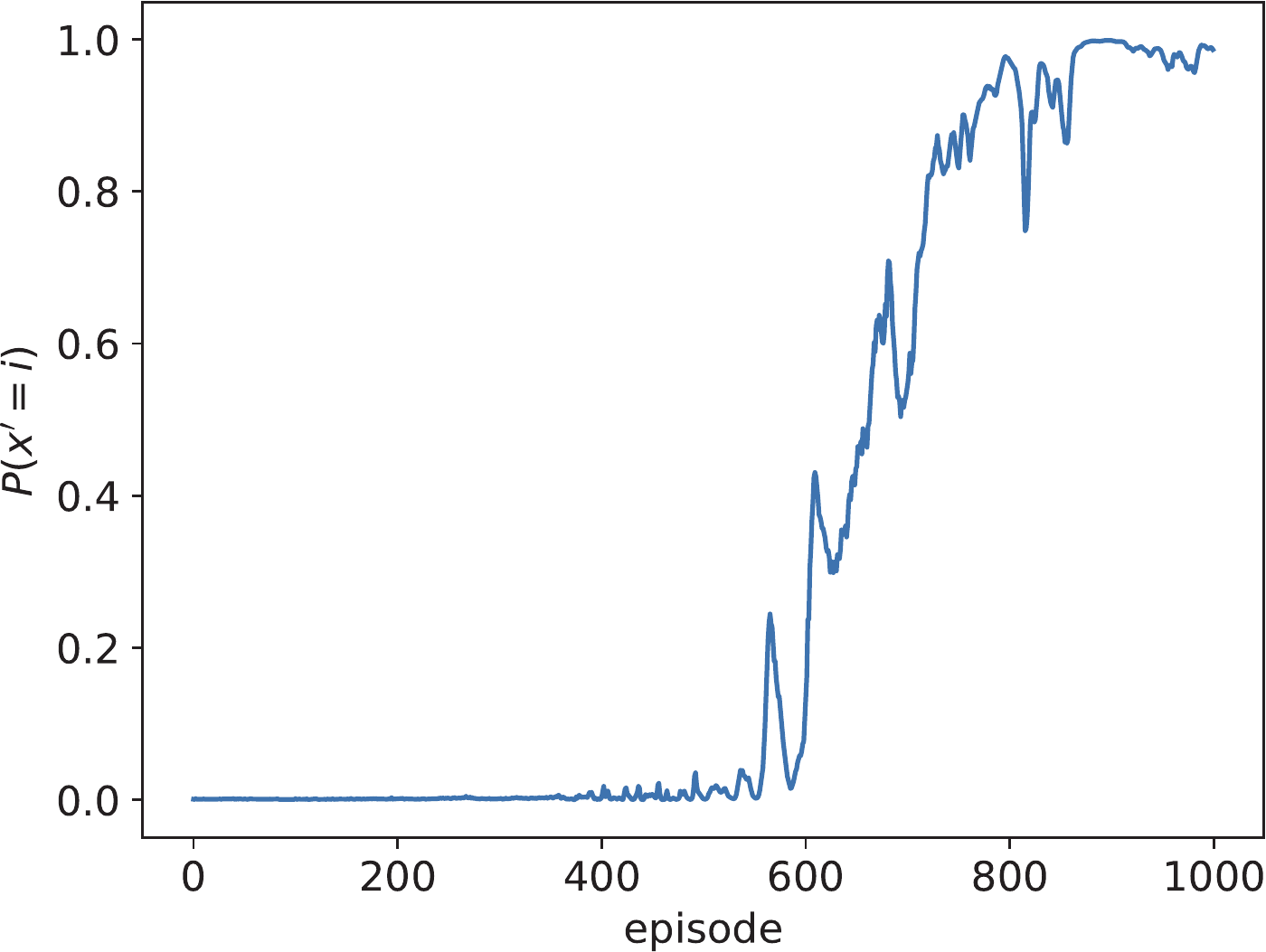}\label{fig:learning_curve}}
		\hspace{20pt}
		\subfigure[Image generated by the actor after $1000$ episodes.]{
			\includegraphics[scale=2]{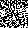}\label{fig:final_image}}
		\caption{Learning results.}
	\end{figure}
	
	In order to generate the fake images, we construct the following simulation model:
	\begin{equation}
	\label{equ:fc}
	f(x)=D\left(r\left(\frac{x+1}{2}\right), i\right),
	\end{equation}
	where $x\in R^{(W\times H)}$ and $r(\cdot)$ is the resizing operator converting the input vector to the appropriate shape. The linear mapping $\frac{x+1}{2}$ converts the value range from $(-1,1)$ to $(0,1)$, which is the same as the images in the date set. Then by solving the optimization problem \eqref{equ:pro}, the actor can generate fake images which can mislead the classifier to believe it as $i$. We choose $i=1$ in our experiments. The learning curve is illustrated in Fig. \ref{fig:learning_curve}, where we can find that within $1000$ episodes (only 1 sample is evaluated by $f(\cdot)$ in each episode), the average confidence reaches over $0.95$. Before the $500$-th episode, the average confidence is close to $0$, which may be caused by the low-accuracy critic. After the $500$-th episode, the score increases gradually, which means both the actor and critic are improved. One of the images generated by the final actor is shown in Fig. \ref{fig:final_image}. It is just some noise from the viewpoint of humans, but the classifier tends to predict it as the digital number ``1''.
	
	Another type of fake images are also preferred, which remain the features of the initial digital number, but mislead the classifier to make wrong decisions by adding some special noise. In order to generate these images, we construct following simulation model:
	\begin{equation}
	\label{equ:in}
	f(x) = D\left(c\left(r\left(\frac{x+1}{2}\right)\times \delta+x_j\right), i\right),~i\neq j
	\end{equation}
	where the operator $c$ clips the input variables to $[0,1]$, and $\delta$ is a small constant controlling the noise level, which is set as $0.2$ in our experiment. Besides, $x_j$ is an image from the initial data set, whose label is $j$. In this case, the simulation-based optimization problem \eqref{equ:pro} tends to find special noise by adding which the classifier will wrongly classify the image from $j$ to $i$.
	
	\begin{figure}
		\centerline{\includegraphics[scale=0.3]{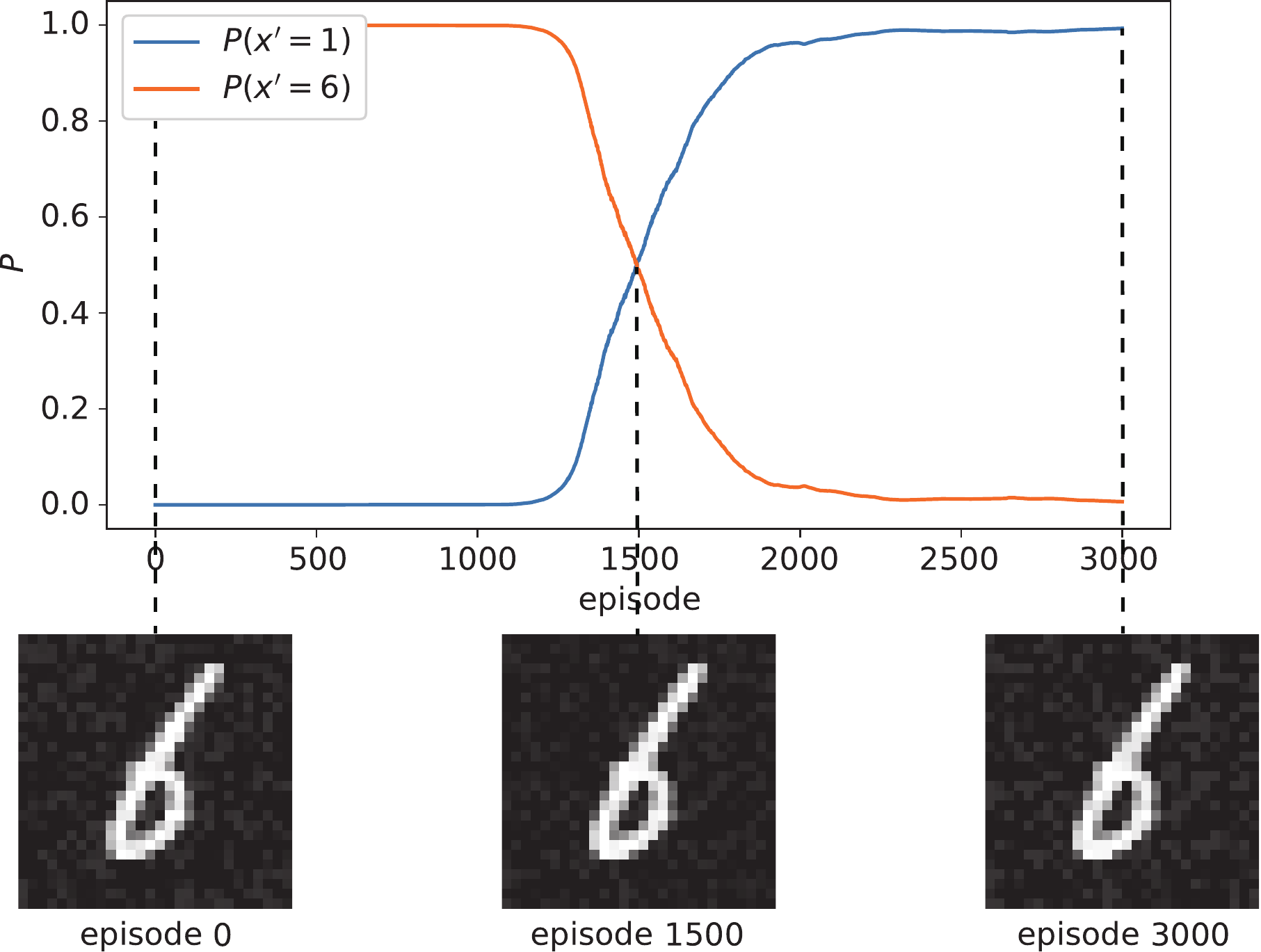}}
		\caption{Result of adding special noise to the initial image. The orange curve is the confidence that the classifier recognizes the noisy image as ``6'', and the blue curve is the confidence of recognizing it as ``1''. The three images in the lower row are the initial image mixed with random noise generated at episode $0$, $1500$ and $3000$, respectively.}
		\label{fig:learning_result}
	\end{figure}
	
	The result is shown in Fig. \ref{fig:learning_result}. After about $2000$ episodes, the actor can generate noise ``special'' enough to mislead the classifier. We present the noisy image at episodes $0$, $1500$, and $3000$. All of these images remain the features of ``6'' in the viewpoint of humans, but the classifier will be ``cheated'' to make wrong decisions.
	
	\begin{figure}
		\centering
		\subfigure[CartPole-v0]{
			\includegraphics[scale=0.4]{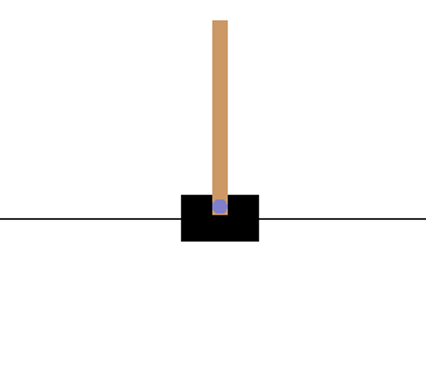}\label{fig:cartpole}}
		\hspace{0pt}
		\subfigure[Learning curve]{
			\includegraphics[scale=0.3]{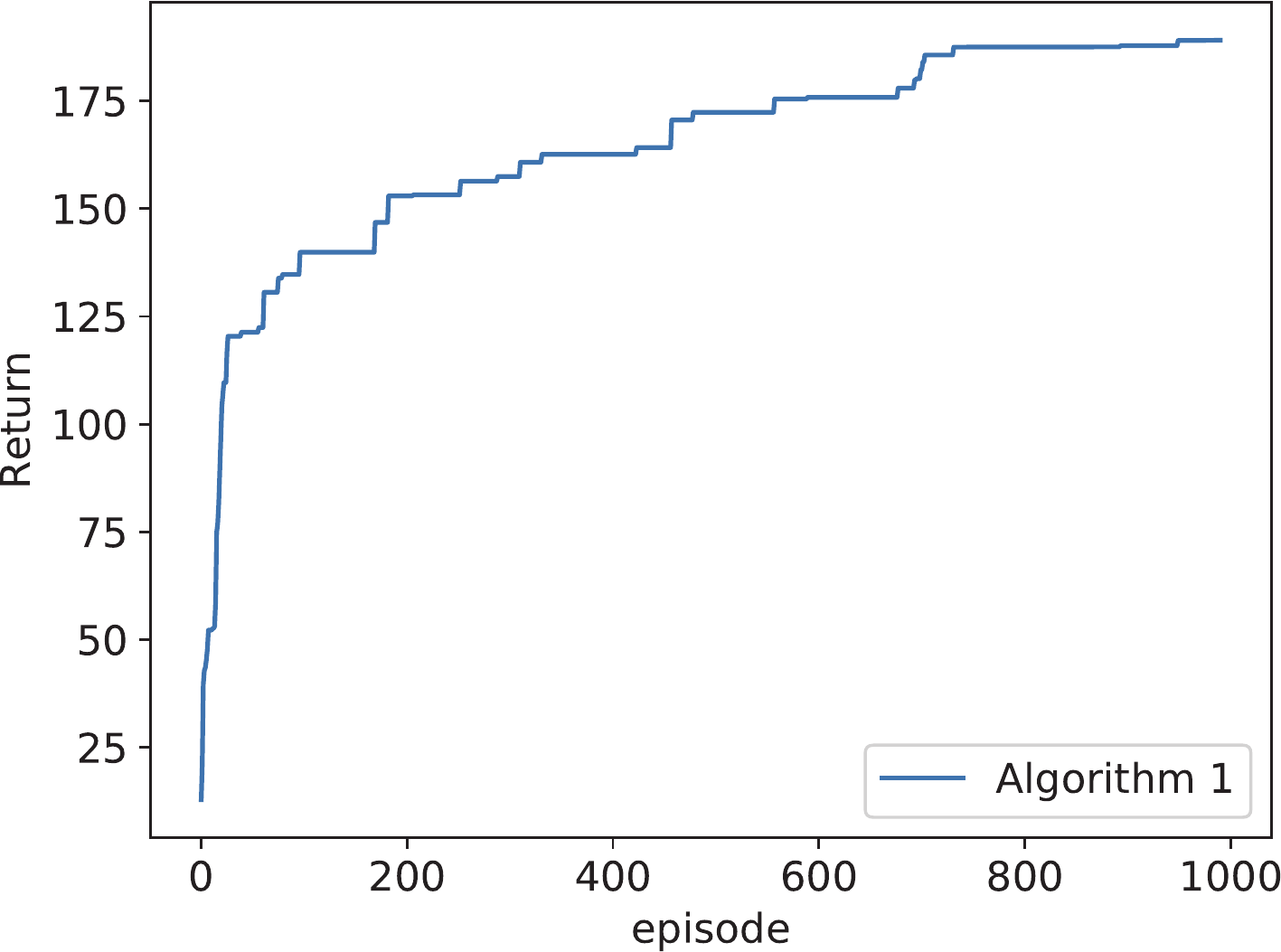}\label{fig:return}}
		\caption{CartPole-v0 environment and the training result. The CartPole-v0 environment takes $0$ or $1$ as action at each time step, which drives the black box toward left or right respectively. At most $200$ steps will be executed in each episode, and the early stop will be triggered if the pole falls down. Each step will be rewarded by $+1$. Fig. \ref{fig:return} shows the learning curves of Algorithm \ref{alg:cvr}, which is drawn by averaging the ``return'' of $300$ independent experiments.}
		\label{fig:rl}
	\end{figure}
	
	Finally, we apply Algorithm \ref{alg:cvr} to a more complex task, i.e., RL task. In typical policy-based and AC-based RL algorithms, the policy is parameterized by $\zeta$ and generates action corresponding to each state. In existing RL methods, the parameter $\zeta$ is adjusted by Policy Gradient (PG) algorithms, as in TRPO (\hspace{1pt}\cite{schulman2015trust}), PPO(\hspace{1pt}\cite{schulman2017proximal}), or Deterministic Policy Gradient (DPG) algorithms, as in DDPG (\hspace{1pt}\cite{lillicrap2015continuous}) and TD3 (\hspace{1pt}\cite{fujimoto2018addressing}), to reach higher accumulated reward. While in this work, we directly treat the task as a simulation-based model, where $\zeta$ is a concrete design. We take the ``CartPole-v0'' environment (\hspace{1pt}\cite{1606.01540}) in Fig. \ref{fig:cartpole} as example and deploy Algorithm \ref{alg:cvr} to it. The result is shown in Fig. \ref{fig:return}, where we can find that the policy (generated design) is successfully reinforced to reach higher score. Although existing methods in RL area have better performance in these RL tasks, our methods offer a new perspective to robot control, which may be helpful to solve the problem of sparse and delayed reward widely existing in RL problems.
	
	We design these examples to show the application of our proposed methods, as well as validate their performance on large-scale problems. The results will be further concluded in the next section.
	
	\section{CONCLUSIONS}
	In this work, we consider the simulation-based optimization problem of selecting the best design with a computationally expensive evaluation model. The sampling process is bridged to the policy optimization problem and solved under the AC framework, where the critic serves as the surrogate model predicting the scores of untested designs, and the actor encodes the sampling distribution. We propose algorithms to update the actor and critic for continuous and discrete design space respectively, and design experiments to validate their effectiveness, as well as explain the applications. The results show that our methods can successfully find satisfying designs and avoid falling into the local optimal to a certain extent. We note that in the experiment of the RL task, we offer a new perspective to robot control of optimizing the policy generating process, rather than optimizing the policy itself. This may be helpful to solve the problem of sparse and delayed reward in existing works. The new perspective will also be studied in future work. Another direction that will be researched in the future is adjusting the hyper-parameter $\alpha$ automatically, as in \cite{haarnoja2018soft2}.
	
	\bibliographystyle{IEEEtran}
	\bibliography{IEEEabrv,references}

\begin{thebibliography}{10}
\providecommand{\url}[1]{#1}
\csname url@rmstyle\endcsname
\providecommand{\newblock}{\relax}
\providecommand{\bibinfo}[2]{#2}
\providecommand\BIBentrySTDinterwordspacing{\spaceskip=0pt\relax}
\providecommand\BIBentryALTinterwordstretchfactor{4}
\providecommand\BIBentryALTinterwordspacing{\spaceskip=\fontdimen2\font plus
\BIBentryALTinterwordstretchfactor\fontdimen3\font minus
  \fontdimen4\font\relax}
\providecommand\BIBforeignlanguage[2]{{%
\expandafter\ifx\csname l@#1\endcsname\relax
\typeout{** WARNING: IEEEtran.bst: No hyphenation pattern has been}%
\typeout{** loaded for the language `#1'. Using the pattern for}%
\typeout{** the default language instead.}%
\else
\language=\csname l@#1\endcsname
\fi
#2}}

\bibitem{boyd2004convex}
S.~Boyd, S.~P. Boyd, and L.~Vandenberghe, \emph{Convex optimization}.\hskip 1em
  plus 0.5em minus 0.4em\relax Cambridge university press, 2004.

\bibitem{BO}
M.~Pelikan, D.~E. Goldberg, E.~Cant{\'u}-Paz, \emph{et~al.}, ``Boa: The
  bayesian optimization algorithm,'' in \emph{Proceedings of the genetic and
  evolutionary computation conference GECCO-99}, vol.~1.\hskip 1em plus 0.5em
  minus 0.4em\relax Citeseer, 1999, pp. 525--532.

\bibitem{GP}
R.~M. Dudley, ``Sample functions of the gaussian process,'' \emph{Selected
  Works of RM Dudley}, pp. 187--224, 2010.

\bibitem{SBOWSC}
Y.~Peng, E.~K. Chong, C.-H. Chen, and M.~C. Fu, ``Ranking and selection as
  stochastic control,'' \emph{IEEE Transactions on Automatic Control}, vol.~63,
  no.~8, pp. 2359--2373, 2018.

\bibitem{lillicrap2015continuous}
T.~P. Lillicrap, J.~J. Hunt, A.~Pritzel, N.~Heess, T.~Erez, Y.~Tassa,
  D.~Silver, and D.~Wierstra, ``Continuous control with deep reinforcement
  learning,'' \emph{arXiv preprint arXiv:1509.02971}, 2015.

\bibitem{fujimoto2018addressing}
S.~Fujimoto, H.~Hoof, and D.~Meger, ``Addressing function approximation error
  in actor-critic methods,'' in \emph{International Conference on Machine
  Learning}.\hskip 1em plus 0.5em minus 0.4em\relax PMLR, 2018, pp. 1587--1596.

\bibitem{haarnoja2018soft}
T.~Haarnoja, A.~Zhou, P.~Abbeel, and S.~Levine, ``Soft actor-critic: Off-policy
  maximum entropy deep reinforcement learning with a stochastic actor,'' in
  \emph{International Conference on Machine Learning}.\hskip 1em plus 0.5em
  minus 0.4em\relax PMLR, 2018, pp. 1861--1870.

\bibitem{DP}
R.~Bellman, ``Dynamic programming,'' \emph{Science}, vol. 153, no. 3731, pp.
  34--37, 1966.

\bibitem{Q-learning}
C.~J. Watkins and P.~Dayan, ``Q-learning,'' \emph{Machine learning}, vol.~8,
  no. 3-4, pp. 279--292, 1992.

\bibitem{haarnoja2017reinforcement}
T.~Haarnoja, H.~Tang, P.~Abbeel, and S.~Levine, ``Reinforcement learning with
  deep energy-based policies,'' in \emph{International Conference on Machine
  Learning}.\hskip 1em plus 0.5em minus 0.4em\relax PMLR, 2017, pp. 1352--1361.

\bibitem{schulman2015trust}
J.~Schulman, S.~Levine, P.~Abbeel, M.~Jordan, and P.~Moritz, ``Trust region
  policy optimization,'' in \emph{International conference on machine
  learning}.\hskip 1em plus 0.5em minus 0.4em\relax PMLR, 2015, pp. 1889--1897.

\bibitem{schulman2017proximal}
J.~Schulman, F.~Wolski, P.~Dhariwal, A.~Radford, and O.~Klimov, ``Proximal
  policy optimization algorithms,'' \emph{arXiv preprint arXiv:1707.06347},
  2017.

\bibitem{1606.01540}
G.~Brockman, V.~Cheung, L.~Pettersson, J.~Schneider, J.~Schulman, J.~Tang, and
  W.~Zaremba, ``Openai gym,'' 2016.

\bibitem{haarnoja2018soft2}
T.~Haarnoja, A.~Zhou, K.~Hartikainen, G.~Tucker, S.~Ha, J.~Tan, V.~Kumar,
  H.~Zhu, A.~Gupta, P.~Abbeel, \emph{et~al.}, ``Soft actor-critic algorithms
  and applications,'' \emph{arXiv preprint arXiv:1812.05905}, 2018.

\end{thebibliography}
\end{document}